%% file: main.tex
\documentclass[letterpaper, 10 pt, journal, twoside]{IEEEtran}

\IEEEoverridecommandlockouts
\usepackage{amsmath,amsfonts}
\usepackage{algorithmic}
\usepackage{algorithm}
\usepackage{array}
\usepackage[caption=false,font=normalsize,labelfont=sf,textfont=sf]{subfig}
\usepackage{textcomp}
\usepackage{stfloats}
\usepackage{url}
\usepackage{verbatim}
\usepackage{graphicx}
\usepackage{cite}
\ifx\argmin\undefined\newcommand{\argmin}{\mathop{\rm argmin}\limits}\fi
\ifx\argmax\undefined\newcommand{\argmax}{\mathop{\rm argmax}\limits}\fi
\ifx\bm\undefined\newcommand{\bm}[1]{\mbox{\boldmath{$#1$}}}\fi
\ifx\um\undefined\newcommand{\um}[1]{{\SI{#1}{\micro \metre}}}\fi %%need \usepackage{siunitx} 

\ifx\etal\undefined\newcommand{\etal}{{\it et al. }}\fi
\ifx\ie\undefined\newcommand{\ie}{{\it i.e.}}\fi
\ifx\eg\undefined\newcommand{\eg}{{\it e.g.}}\fi
\ifx\gt\undefined\newcommand{\gt}{$\textgreater$}\fi
\ifx\lt\undefined\newcommand{\lt}{$textless$}\fi

\title{TIDE: Temporally Incremental Disparity Estimation via Pattern Flow in Structured Light System}

\author{Rukun Qiao$^{1}$, Hiroshi Kawasaki$^{2}$ and Hongbin Zha$^{1}$
\thanks{Manuscript received: October, 26, 2021; Revised January, 7, 2022; Accepted Januaray, 28, 2022.}
\thanks{This paper was recommended for publication by Editor Sven Behnke upon evaluation of the Associate Editor and Reviewers' comments.}
\thanks{$^{1}$Rukun Qiao and Hongbin Zha are with the Key Lab of Machine Perception (MOE), School of Artificial Intelligence, Peking University, Beijing, China.
        {\tt\small rukunqiao@pku.edu.cn, zha@cis.pku.edu.cn}}%
\thanks{$^{2}$Hiroshi Kawasaki is with the Graduate School and Faculty of Information Science and Electrical Engineering, Kyushu University, Fukuoka, Japan.
        {\tt\small kawasaki@ait.kyushu-u.ac.jp}}%
\thanks{Digital Object Identifier (DOI): 10.1109/LRA.2022.3150029. Copyright:~\copyright~2022 IEEE.}
}

\begin{document}

\maketitle

% \IEEEpubidadjcol

\begin{abstract}

We introduced Temporally Incremental Disparity Estimation Network (TIDE-Net), a learning-based technique for disparity computation in mono-camera structured light systems. In our hardware setting, a static pattern is projected onto a dynamic scene and captured by a monocular camera. Different from most former disparity estimation methods that operate in a frame-wise manner, our network acquires disparity maps in a temporally incremental way. Specifically, We exploit the deformation of projected patterns (named \emph{pattern flow}) on captured image sequences, to model the temporal information. Notably, this newly proposed pattern flow formulation reflects the disparity changes along the epipolar line, which is a special form of optical flow. Tailored for pattern flow, the TIDE-Net, a recurrent architecture, is proposed and implemented. For each incoming frame, our model fuses correlation volumes (from current frame) and disparity (from former frame) warped by pattern flow. From fused features, the final stage of TIDE-Net estimates the residual disparity rather than the full disparity, as conducted by many previous methods. Interestingly, this design brings clear empirical advantages in terms of efficiency and generalization ability. Using only synthetic data for training, our extensitve evaluation results (w.r.t. both accuracy and efficienty metrics) show superior performance than several SOTA models on unseen real data. The code will be available on \textit{https://github.com/CodePointer/TIDENet} soon.

\end{abstract}


\input{chapters/1_Introduction.tex}

\input{chapters/2_RelatedWork.tex}

\input{chapters/3_Methods.tex}

\input{chapters/4_Experiments.tex}

\input{chapters/5_Conclusion.tex}

{
    \bibliographystyle{IEEEtran}
    \bibliography{egbib}
}

\end{document}

%% file: chapters/1_Introduction.tex
\section{Introduction}

% Para1: 
\IEEEPARstart{V}{arious} variants of disparity estimation methods built upon monocular structured light systems~\cite{salvi2004pattern, salvi2010a} have been proposed in the literature. They have drawn wide attention from both academia and industry due to their promises in important application areas like augmented reality, sport analysis, and medical robots. However, dynamic scene acquisition is still an unsolved challenging problem. While object scanning in static scenes can leverage rich information provided by multiple patterns, this setting is not suitable for dynamic scenes due to the difficulty of feature matching. As such, for dynamic scenes, single-pattern structured light systems are preferred, which is refereed to as \emph{one-shot scan} in the literature. Since input sequences in the one-shot scan setting tend to be sparse and unstable, recently several learning-based methods are proposed and widely recognized as a promising alternative solution to address these two challenges~\cite{ryan2016hyperdepth, fanello2017ultrastereo, riegler2019connecting}.

% Para2:
Although these deep neural network (DNN) models have achieved promising accuracy, there are two obstacles preventing us from training robust models with large-scale real-world datasets: 1. getting the ground truth of dense correspondences of dynamic scenes is virtually impossible; 2. even if we could get these ground truths, the data would be biased towards the specific pattern and/or device in the hardware design. Therefore, training on the synthetic data and testing on real-world scenarios is a more practical choice. In such a situation, the generalization ability of the model is the most significant factor to consider. 

% Para3:
In this paper, the focus is to exploit the temporal coherence in the image sequence to improve the generalization ability, while guaranteeing efficiency. To this end, we propose a neural network named TIDE-Net (\textbf{T}emporally \textbf{I}ncremental \textbf{D}isparity \textbf{E}stimation). Rather than estimating a disparity map for each frame from scratch, TIDE-Net conducts residual estimation based upon the former frame. Notably our method can model even earlier frames, through hidden layers in a recurrent architecture. This incrementally updated scheme allows the network size to be concise while maintaining accuracy.

% Para4
Utilizing temporal coherence between frames in a video has been studied intensively for correspondence search~\cite{taguchi2012motion-aware, riegler2019connecting}. However, the naive practice of aggregating features extracted from frames by concatenation ignores the pixel displacement caused by scene motion, and thus, it usually results in low accuracy~\cite{hosni2011temporally}. To address this issue, we propose to leverage a novel formulation named \emph{pattern flow}, which is the pattern deformation in the observed image and a constrained version of generic optical flow. Although a former study has mentioned the pattern flow concept~\cite{furukawa2017depth}, they treat it as a feature instead of explicitly using the pattern flow for correspondence search between a captured image and a projected pattern.

% Para5
In contrast, we analyzed the pattern flow and found that it has a unique feature absent in the generic optical flow formulation: The disparity changes between frames can be computed explicitly given the estimated pattern flow. Once the pattern flow is estimated, we can pixel-wisely warp both disparity and hidden layer from the former frame. Thus, features and information can be aligned along the temporal space credit to this operation. Experimental results illustrated that our method required less computational cost than SOTA methods while guaranteeing better accuracy and generalization ability.

% Para6:
In summary, our contributions are threefold: 
\begin{itemize}
  \item We propose an incremental disparity estimation framework based on TIDE-Net which fully utilized the local and sequential nature of images from dynamic scenes. By focusing on the incremental non-linear portion, parameter size can be reduced while guaranteeing accuracy and efficiency.
  \item We propose a novel algorithm to estimate pattern flow, which represents correspondences of projected pattern between adjacent frames of structured light systems, and the information is fed into TIDE-Net.
  \item Comprehensive experiments proved that, although TIDE-Net is only trained by synthetic data and evaluated by real one without any adaptation, our method performs better accuracy than several SOTA models as well as computational cost, showing efficient domain-invariant generalization ability of the method.
\end{itemize}

%% file: chapters/2_RelatedWork.tex
\section{Related Work}\label{chap:relatedwork}

The active light systems can achieve robust and accurate depth information even if there is no texture or high-frequency shape on the object surface. Thus, they are considered a practical solution for 3D scan~\cite{salvi2004pattern, salvi2010a}. For reconstruction on dynamic scenes, many spatial coding techniques have been proposed based on pattern decoding and global matching~\cite{zhang2002rapid, kawasaki2015active, furukawa2016shape, furukawa2017depth}. The pattern can be binary dots~\cite{vuylsteke1990range}, strips~\cite{kawasaki2008dynamic}, grids~\cite{proesmans1996one}, etc. These methods focus on encoding coordinate features into one projected pattern, and then extracting features from observed images by decoding on the spatial domain. The projected pattern needs to be recognizable in the images; thus, the computed disparity map is always sparse. 

As for the utilization of temporal information, some methods embedded spatial features in the multiple different patterns projected periodically. To compensate moving parts in the scene for a one-shot scan, several methods are proposed, such as detecting individual motion~\cite{taguchi2012motion-aware, zhang2014realtime}, directly computing disparity of the scene by the blurs of line-based patterns~\cite{furukawa2017depth}, or utilizing DNN~\cite{riegler2019connecting}. In the method~\cite{riegler2019connecting}, they propose a geometric loss between image pairs for their model training to improve the spatio-temporal consistency. The loss requires objects to have a rigid motion, and related position $(R, t)$ is needed for model training. Our research does not have rigid motion assumptions, and no registration is needed for training. 

Stereo matching and optical flow focusing on the correspondences between image pairs are also related to our problem since a pattern projector is optically the same as a camera. Many networks designed for those problems can be applied to structured light systems with several additional modifications~\cite{zhang2018active, zbontar2016stereo, teed2020raft}. We exploit ideas from the stereo methods to our rectified mono-camera setup, where the reference pattern is warped to the observed image by correlation layer for disparity estimation inside the network structure.

The domain generalization ability of networks is also an important issue and drawing wide attention recently. Some methods apply self-supervised training~\cite{riegler2019connecting, johari2021depthinspace} or fine-tuning when transfer to a new data domain. However, such solutions require a large number of data from the target domain with or without ground truth. Some methods apply online adaptation techniques to update the network during the inference process~\cite{tonioni2019realtime, li2020selfsupervised}, while the BP process results in the loss of efficiency. \cite{zhang2019domaininvariant} proposed DSMNet equipped with a novel normalization layer and graph-based filtering layer to improve the generalization ability for passive stereo matching. In contrast, we do not design a specific framework in our network. We utilize the temporal coherence from the input sequence to guarantee the accuracy and efficiency of our network.

%% file: chapters/3_Methods.tex
\begin{figure}[thpb]
    \centering
    \framebox{\includegraphics[width=\linewidth]{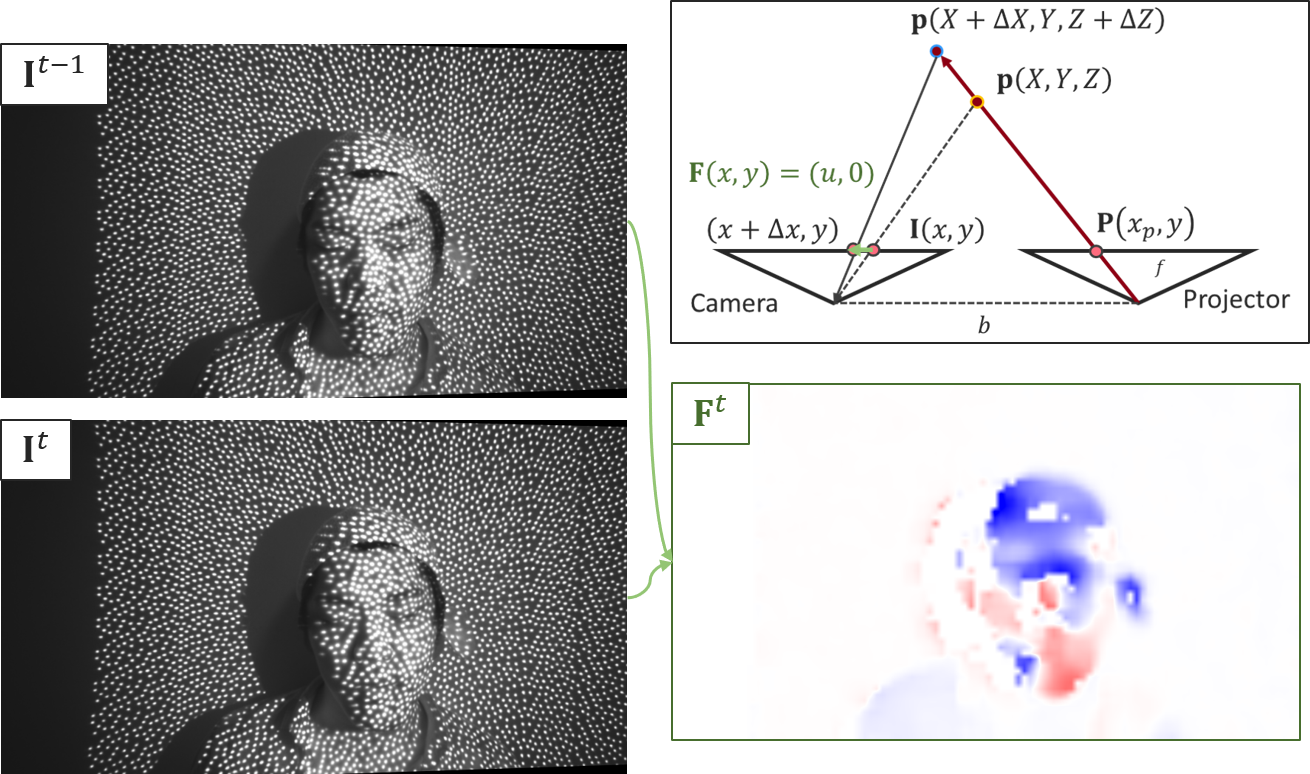}}
    \caption{{\it Pattern flow} from adjacent image pairs. We denote red and blue colors for different moving directions, while intensity stands for the value. The forehead is moving back and the jaw is moving forward. Top-right is a sketch map for the geometric property of the pattern flow in structured light systems.}
    \label{fig:pfcase}
\end{figure}

\begin{figure*}[thpb]
    \centering
    \framebox{\includegraphics[width=0.9\textwidth]{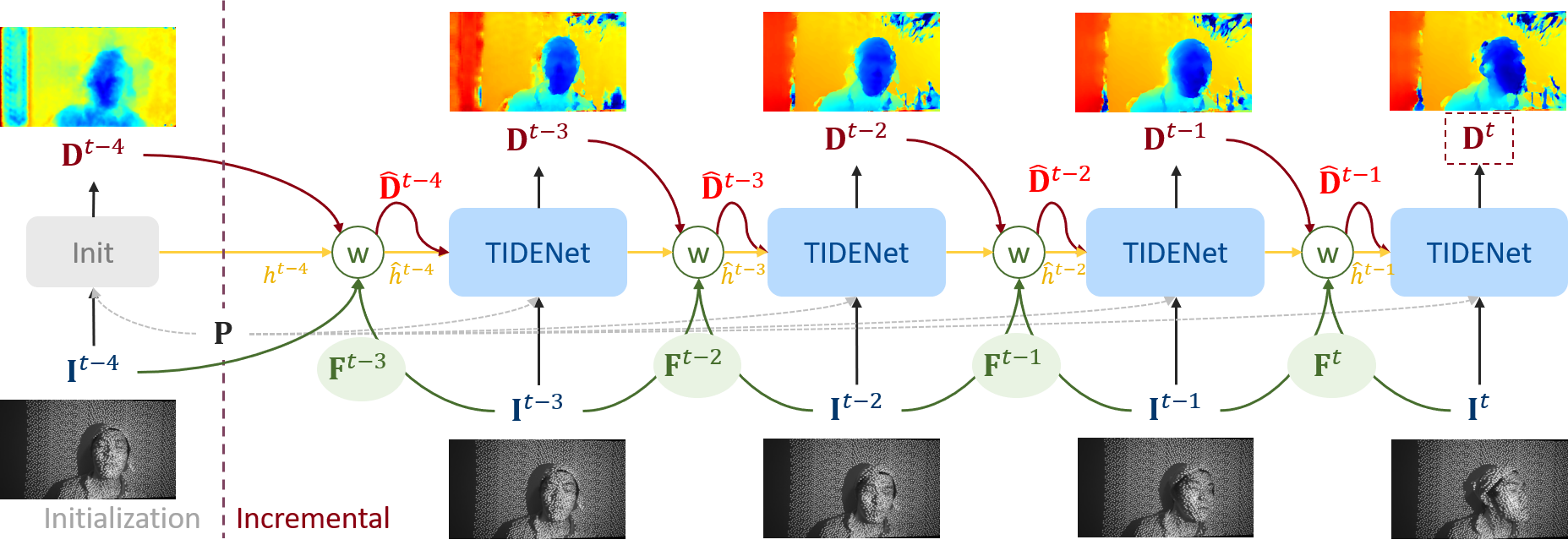}}
    \caption{The general framework of our method. The initialization process is only applied for the first frame, then we incrementally estimate the disparity map for every frame using TIDE-Net.}
    \label{fig:framework}
\end{figure*}

\begin{figure}[thpb]
    \centering
    \framebox{\includegraphics[width=\linewidth]{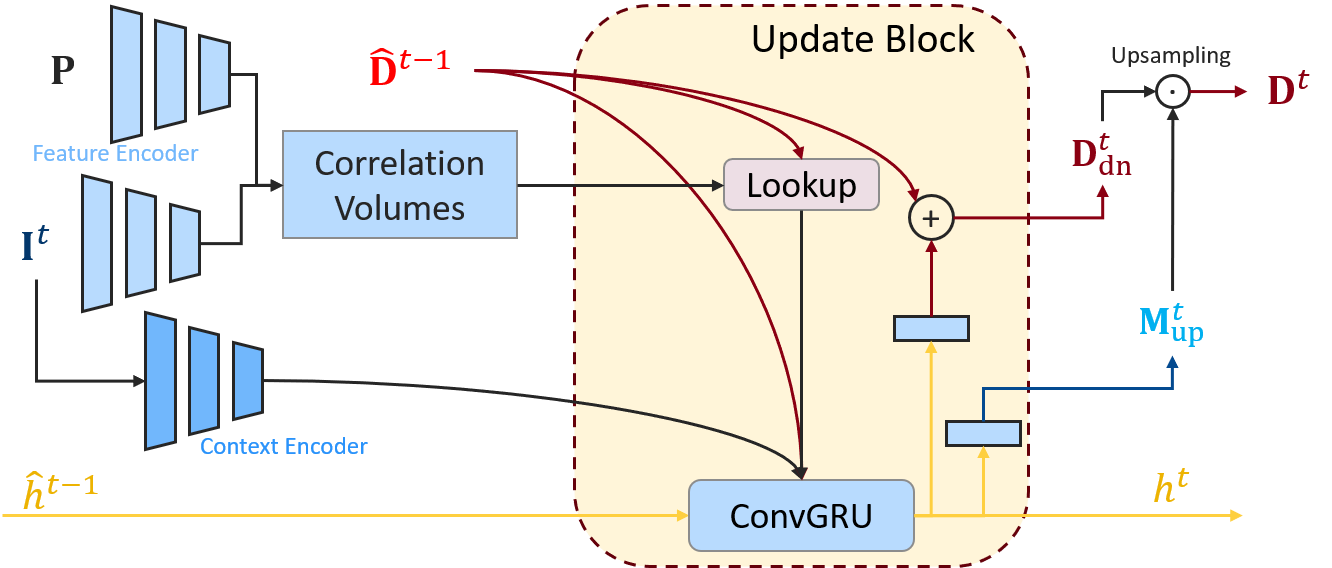}}
    \caption{The architecture of the TIDE-Net.}
    \label{fig:network}
\end{figure}

\begin{table}
	\begin{center}
	\begin{tabular}{|l|c|c|}
			\hline
			& Optical Flow & Pattern Flow           \\
            \hline
            Definition & 3D point & Light ray \\
            \hline
			Degree of freedoms & 2 & 1               \\
			Physical correspondence  & Yes & No              \\
			Texture-less problem & Yes & No \\
			Intensity model & Passive & Active \\
			\hline
		\end{tabular}
	\end{center}
	\caption{The comparison between classic optical flow and proposed pattern flow.}
	\label{tab:compare}
\end{table}

\section{Methods}\label{chap:methods}
In this part, we will illustrate our method in detail. We first go through the definition of incremental disparity computing from an image sequence in structured light systems. Then, we explain the definition and analysis of pattern flow. Finally, we introduce the entire framework combined with the incremental warping technique.

\subsection{Problem Definition}\label{subsec:definition}
Our research aims at the 3D depth estimation from sequential frames where scenes are projected by a static pattern projector and captured by a single camera. For such systems, the pattern projector can be regarded as a second camera considering its image plane as the reference pattern. Therefore, we can define disparity for the camera similar to stereo vision accordingly. We assume that the camera and the projector are calibrated and rectified by pre-processing. Generally, scenes are captured by the camera, and captured images are fed into the algorithm to estimate disparity maps for each frame. Here, we denote $\mathrm{\mathbf{I}}^t$ as the input camera image and $\mathrm{\mathbf{D}}^t$ as the output disparity, where $t$ represents the frame number. The designed pattern is denoted as $\mathrm{\mathbf{P}}$. In our system, a pseudo-random dot pattern~\cite{realsense} is used, which guarantees patch uniqueness in each row and is repeated periodically to the vertical direction.

We assume online disparity estimation for our system, that is, for each input frame $\mathrm{\mathbf{I}}^t$, all the history information before time $t$ is available for calculation, denoted as $H^{t-1}$. Thus, the disparity of the current frame can be estimated through incremental processing:

\begin{equation}\label{formula:rawdef}
	\mathrm{\mathbf{D}}^t = \mathrm{\mathbf{D}}^{t-1} + 
	g\left(
	    \mathrm{\mathbf{I}}^t, \mathrm{\mathbf{P}}, H^{t-1}
	\right),
\end{equation}

where we assume $t > 0$. Our incremental process, denoted as function $g$, focuses on the residual estimation instead of the full disparity computation for each frame. This usually helps to improve accuracy and efficiency. Rather than some method utilizing temporal information as an additional dimension for temporal-window-based batch matching \cite{zhang2003spacetime}, our method focuses on the step-by-step updating processing. We utilize a deep neural network (DNN) model with a temporal module to find the function $g$, where the hidden layer is passed through the sequence as the history information $H$. This is a straight-forward idea and has been developed in stereo matching problems.

However, simply concatenating multiple frames usually ignores the motion of objects in the scene, resulting in the overfitting on specific motion features in the synthetic dataset. To achieve the network to learn the motion-invariant consistent features explicitly, we pre-warp the history information consisting of previous frames to retrieve accurate correspondences between adjacent frames. We proposed to use pattern flow between frames to accomplish the warping. The pattern flow can be robustly calculated locally with a simple but effective algorithm with our proposed method. From the ablation study in Sec. \ref{chap:experiments}, we can find that the pattern flow can improve the performance compared to simply passing the hidden layer directly.

\subsection{Pattern Flow in Structured Light Systems}\label{sec:patternflow}
In a structured light system, the projector emits a reference pattern into the 3D space. The projected pattern is deformed by the shape of the object's surface and observed by the camera. If objects in the scene are moving, the observed patterns are deformed, because the shape (depth) of the scene also changes as shown in Fig.~\ref{fig:pfcase}(left), where two adjacent frames are shown to visualize the motion of the projected pattern. We define this kind of light flow caused by projected pattern as {\it pattern flow}, denoted as $\mathrm{\mathbf{F}}^t$, which represents the motion of the projected patterns. This concept was mentioned by Furukawa \etal, who use strip pattern flow to estimate depth for fast-moving object~\cite{furukawa2017depth}. In our research, we put forward the pattern flow defined on the pixel level to propagate information precisely. 

We first give out the geometric relationship between the disparity map and the pattern flow. Considering a light ray projected from $\mathrm{\mathbf{P}}(x_p, y)$ on reference pattern to the 3D position $\mathbf{p}(X,Y,Z)$ at time $t$. In the next frame, the projected position moves to a new place caused by the scene motion, denoted as $\mathbf{p}(X+\Delta X,Y,Z+\Delta Z)$. Notice that these two positions are not guaranteed to be the same physical point, but they are on the same light ray projected from the projector. Therefore the $\mathbf{p}$ can only move along the epipolar plane, and we do not need to consider vertical direction on the rectified image. We denote $f$ as the focal length, $b$ as the baseline, and $(x, y)$ as the observed pixel in camera space. Then we have
\begin{equation}\label{formula:projection}
    \left\{
        \begin{split}
             & x = \frac{f}{Z}X \\
             & x_p = \frac{f}{Z}(X - b)
        \end{split}
    \right. .
\end{equation}
By taking the derivative of Eq.\ref{formula:projection}, we have
\begin{equation}\label{formula:dprojection}
    \left\{
        \begin{split}
             \frac{\mathrm{d}x}{\mathrm{d}t} 
             & = \frac{f}{Z}\frac{\mathrm{d}X}{\mathrm{d}t} 
             - \frac{fX}{Z^2}\frac{\mathrm{d}Z}{\mathrm{d}t} \\
             \frac{\mathrm{d}x_p}{\mathrm{d}t} 
             & = \frac{f}{Z}\frac{\mathrm{d}X}{\mathrm{d}t} 
             - \frac{f(X-b)}{Z^2}\frac{\mathrm{d}Z}{\mathrm{d}t}
        \end{split}
    \right. .
\end{equation}

Since the two projected positions belong to the same projected ray, we have $\frac{\mathrm{d}x_p}{\mathrm{d}t}=0$. The pattern flow can also be denoted as a horizontal vector $\mathrm{\mathbf{F}}(x, y)=(u,0)$, where $u = \frac{\mathrm{d}x}{\mathrm{d}t}$. The Eq.~\ref{formula:dprojection} can be further simplified as
\begin{equation}\label{formula:u}
    u = - \frac{fb}{Z^2}\frac{\mathrm{d}Z}{\mathrm{d}t}.
\end{equation}

On the other hand, suppose $d$ is the disparity value for pixel $\mathrm{\mathbf{D}}(x, y)$. Therefore we have $d = \frac{fb}{Z}$ according to the epipolar geometry. Take the derivative of $d$ we have
\begin{equation}\label{formula:d}
    \frac{\mathrm{d}d}{\mathrm{d}t} = - \frac{fb}{Z^2}\frac{\mathrm{d}Z}{\mathrm{d}t}.
\end{equation}
Thus from Eq.~\ref{formula:u}, \ref{formula:d} we can have:
\begin{equation}
    u = \frac{\mathrm{d}d}{\mathrm{d}t},
\end{equation}
\textbf{which means that the estimated {\it pattern flow} is the change of the disparity between images.} The geometric relationship is shown in Fig~\ref{fig:pfcase}(top-right) for better understanding. With the help of pattern flow, the disparity can be propagated between frames according to Eq.\ref{formula:d}. In our framework, we propagate both disparity map and hidden layer to eliminate the motion factor between frames.

In structured light systems, the intensity is decided by the projected ray and object texture together. However, in our practice, the projected light always holds a dominant position in IR image in most cases. Therefore, the local and fast Lucas-Kanade method along the epipolar line could be applied for pattern flow calculation. 

Notice that although we apply a classical method from optical flow calculation, the pattern flow has several differences to classic optical flow in the passive sensor:

\begin{itemize}
  \item Firstly, the optical flow is the motion for physical points, while the pattern flow is the displacement of projected light rays, which means it only appears along the 1D epipolar line, rather than the 2D image. 
  
  \item Secondly, pattern flow in active light is always easier to calculate, for the pattern always has enough features for local matching. No semi-global optimization is needed for dense pattern flow calculation.
  
  \item Lastly, although the pattern flow has the before-mentioned advantage to optical flow, it does not guarantee the tracked points in two frames belong to the same physical points. Pattern flow lacks physical correspondence information and cannot be used for motion estimation. However, it can be used for disparity propagation from the previous disparity.
\end{itemize}

The differences between pattern flow and optical flow are listed in Table \ref{tab:compare} for better illustration.

\subsection{TIDE Network}

Given an observed image sequence clip $\mathrm{\mathbf{I}^{t-k:t}}$ from the camera and pre-designed pattern $\mathrm{\mathbf{P}}$, we estimate the disparity sequence with iterative updating process frame by frame. Fig.~\ref{fig:framework} shows an overview of our approach, and the network architecture is shown in Fig.~\ref{fig:network}. We now describe all the components in detail.

\subsubsection{Feature extraction}
We concatenate the input image $\mathrm{\mathbf{I}}^t$ and the normalized image from local contrast normalization(LCN) method~\cite{zhang2018active, riegler2019connecting} as the input of feature extraction. For each frame inside the temporal window, two encoders with the same architecture are applied to the input image $\mathrm{\mathbf{I}}^t$ and pattern $\mathrm{\mathbf{P}}$. We use the similar feature extraction layers mentioned in ~\cite{teed2020raft} but reduce the parameter numbers. Then, the correlation pyramid is built which will be indexed by updating block. 

\subsubsection{Update block}
We use the similar recurrent updating block in \cite{teed2020raft} for disparity refinement, where a ConvGRU block is applied to provide disparity residual estimation. The input of ConvGRU block is concatenated from 3 parts: The warped disparity $\mathrm{\mathbf{\hat{D}}}^{t-1}$, the feature from the context encoder, and the retrieved correlated features from the correlation pyramid given the predicted disparity. The ConvGRU block outputs disparity at 1/8 resolution. We then upsample the disparity map to full resolution by taking an upsampling mask as the interpolation weights. Although the network used in \cite{teed2020raft} can acquire accurate results on optical flow estimation, it is utilized between image pairs, resulting in the lack of temporal information usage. Besides, the infer speed is also harmed by multiple recurrent iterations. Thus, we improve the framework in two main parts: 1. We passed the hidden layer along the different frames to provide temporal consistency; 2. Instead of taking multiple iterations in every frame, we only have one iteration for each frame, forcing the network to get information from the hidden layer. With those improvements, the computing time could be reduced while guaranteeing accuracy.

\subsubsection{Warping by pattern flow}
As mentioned at the end of Sec.\ref{subsec:definition}, simply passing the previous hidden layer and disparity map may result in the bias by objects motion. Therefore, we first calculate the pattern flow $\mathrm{\mathbf{F}}^t$ between $\mathrm{\mathbf{I}}^t$ and $\mathrm{\mathbf{I}}^{t-1}$, then warp the hidden layer $h^{t-1}$ and disparity map $\mathrm{\mathbf{D}}^{t-1}$ (The "w" operator in Fig.\ref{fig:framework} between frames). We calculate the pattern flow by applying Lucas–Kanade method for every pixel in $1/8$ resolution, which is the input resolution for the update block. Thanks to the feature-rich projected pattern, the pattern flow can be calculated without any semi-global optimization. The dense and accurate pattern flow map could be acquired in a very short time.

\subsubsection{Initialization for the first frame}
Since our method is an incremental function, we need the initial value of the first frame. In practice, we found a very rough disparity estimation is enough for TIDE-Net to reconstruct the accurate disparity within several frames. Thus, we train a small U-Net~\cite{riegler2019connecting} to give the initial output for the very first frame. The initialization process is only needed for the very first frame of the image sequence.

\subsection{Loss Functions}
We applied L1 loss for supervised training on the synthetic dataset given the disparity ground truth:

\begin{equation}
    L = \sum_{i=t-N+1}^{t} \left|
        \mathrm{\mathbf{D}}^{i} - \mathrm{\mathbf{D}}^{i}_{GT}
    \right|
\end{equation}

where the $N$ denotes the temporal window size for the training.

%% file: chapters/4_Experiments.tex
\begin{figure}[thpb]
    \centering
    \framebox{\includegraphics[width=\linewidth]{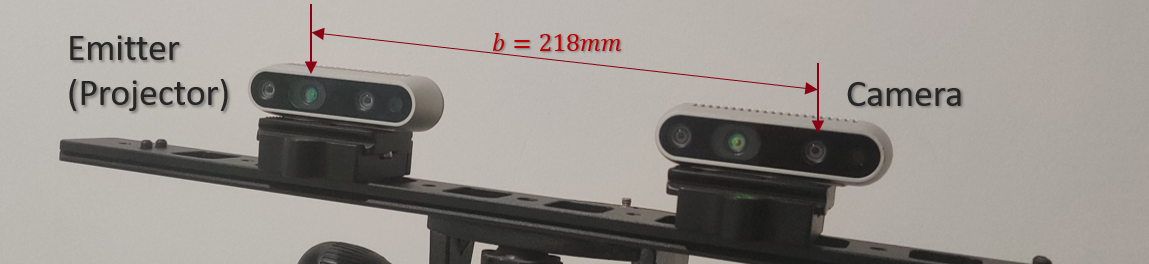}}
    \caption{The experiment platform for data collection.}
    \label{fig:device}
\end{figure}

\begin{figure*}[thpb]
    \centering
    \framebox{\includegraphics[width=0.9\linewidth]{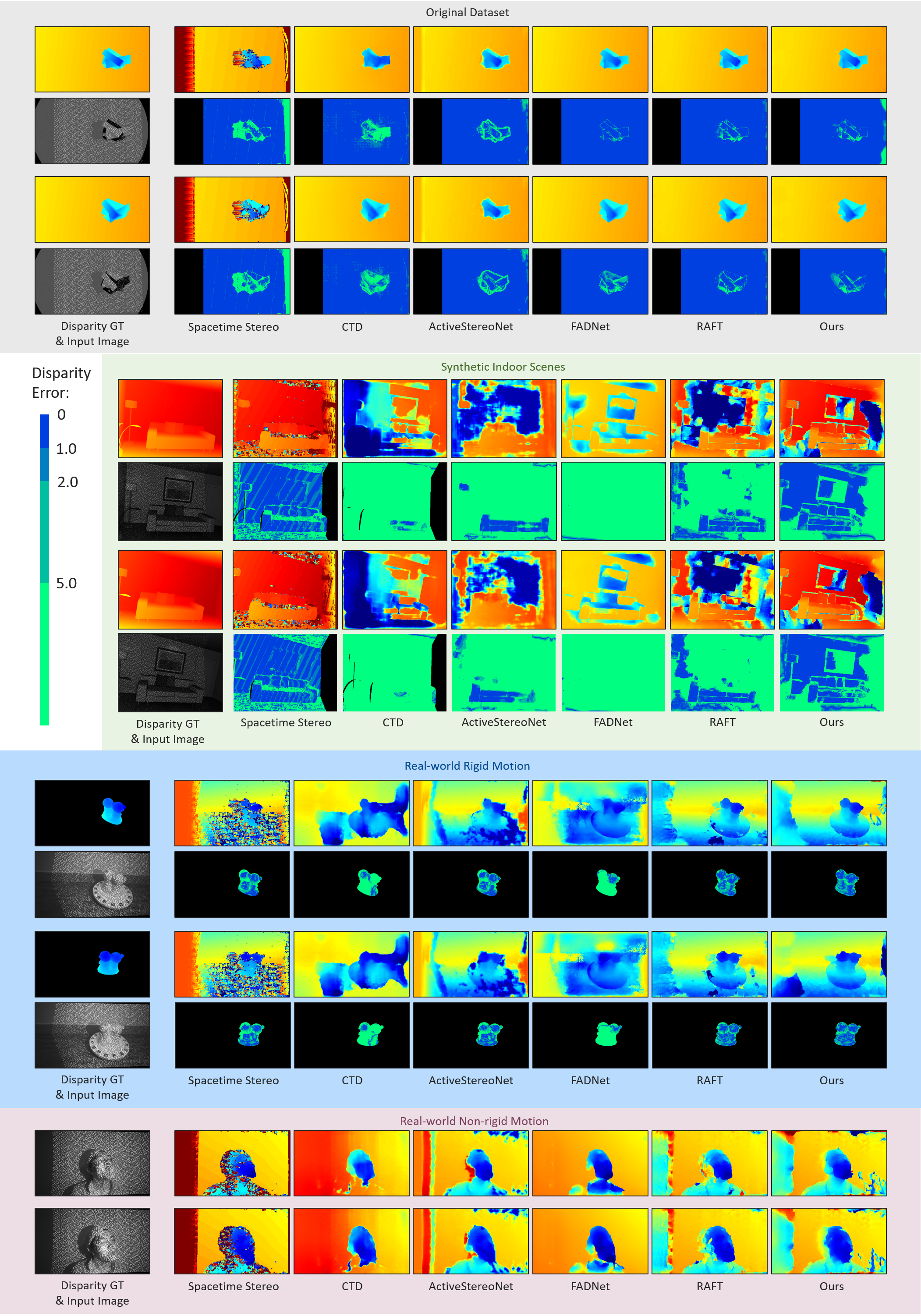}}
    \caption{The experiment results for visualization. All the models are trained on the original synthetic data, the first row, then evaluated on the others. We show the disparity map (upper row) and residual error map(lower row) for comparison. The non-projected area is masked out manually and excluded from the quantitative analysis.}
    \label{fig:realviz}
\end{figure*}

\begin{figure*}[thpb]
    \centering
    \framebox{\includegraphics[width=0.9\linewidth]{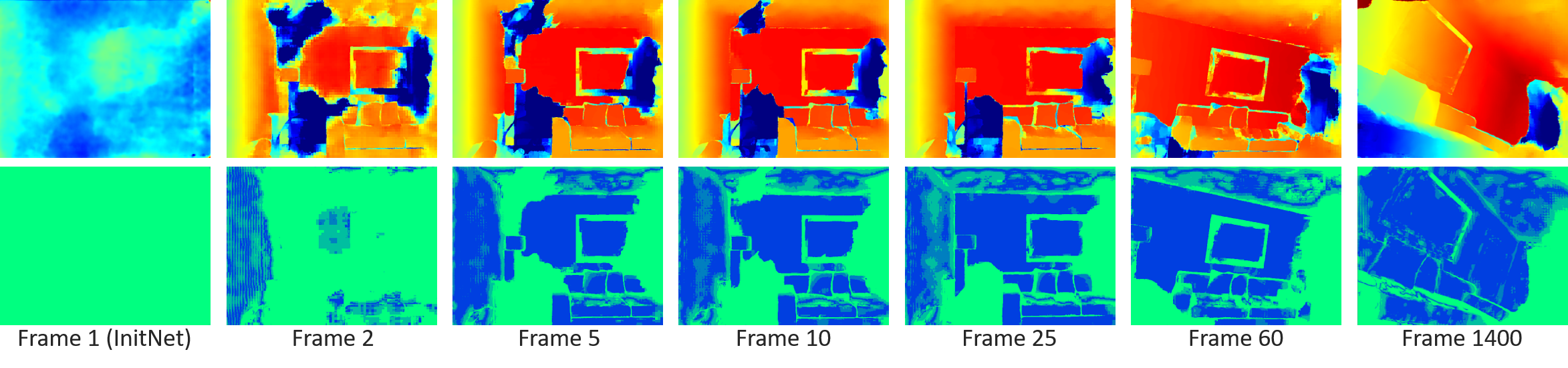}}
    \caption{Our experiment result on the synthetic sequence. With several processing frames(about 30-50), our method can achieve an accurate result.}
    \label{fig:seqviz}
\end{figure*}

\section{Experiments}\label{chap:experiments}

In this section, we systematically evaluate our method compared with the previous technique. We first introduce the experiment platform we used for data collection, then we give out the details of the training process. Finally, we present the evaluation result on several unseen data to illustrate our method's generalization ability and the contribution of temporal information.

\subsection{Experiment Platform}
We use the realsense D435i\cite{realsense}, an active stereo system equipped with two infrared cameras and one emitter, to collect real data for evaluation. In our research, we only utilize one camera for evaluation. We also manually increase the baseline to 218mm by utilizing another D435i device, as shown in Fig.~\ref{fig:device}, for we find the original baseline is not sensitive enough to the object motion. We combine one camera from the left D435i and the emitter from another as the final mono-camera structured light system. The device collects and stores the images with fps=24.

The pattern we used for our experiments is the pseudo-random dots that were originally applied in realsense D435i. Since they do not provide the pre-designed pattern officially, we calibrate the pattern by moving the device forward and backward towards a white wall. Then we apply rectification to the camera image and calibrated pattern to get the data for evaluation. We project that pattern in the scene and captured the images with the left IR camera. All the methods take the IR image sequence as the network input.

\subsection{Training Schedule and Implementation Details}\label{subsec:dataset}
Once we calibrated the devices used for data collection, we can generate the synthetic dataset for model training and then evaluate that on the real collected data. To generate image sequences of dynamic scenes with non-rigid motion, we randomly sample 4 objects from ShapeNet~\cite{chang2015shapenet}, each of which has an individual random rigid motion. The camera and projector are placed towards those objects, with a white plane as the background. Then, we utilized the structured light renderer provided in ~\cite{riegler2019connecting} for infrared image generation with the pre-calibrated parameters used for real data collecting. With those settings, we generate 2048 sequences with 32 frames for each. We also generate another 512 sequences for the test, shown in Fig.~\ref{fig:realviz} (Original Dataset).

For model training, we use Adam optimizer with the learning rate at $10^{-3}$. We trained our model on one 2080Ti GPU with the batch size $N=1$. The temporal window size is set to 8, for this is the maximum capacity of the GPU memory.

\subsection{Qualitative Result on Real Data}
For non-rigid data evaluation, we take a white wall as the background, and collect the images of the non-rigid moving human face. We select several SOTA methods in stereo matching and optical flow for comparison with some small modifications to structured light systems: 1. The Active Stereo Net~\cite{zhang2018active} for active stereo; 2. the FADNet~\cite{wang2020afast} for passive stereo matching; 3. the RAFT~\cite{teed2020raft} framework for optical flow, and 4. CTD~\cite{riegler2019connecting} framework using temporal information for structured light systems. Apart from those deep-learning methods, we also add the Spacetime stereo~\cite{zhang2003spacetime} for comparison. All the comparison baselines take single frame as input during the evaluation while ours takes multi-frames for incremental estimation. With the help of temporal information, ours can perform especially better on the occluded part with a better prediction from history, as shown in Fig.~\ref{fig:realviz} (Real-world Non-rigid Motion).

\subsection{Quantitative Result}
To better illustrated our model's performance, we test our method on several data with the given ground-truth for quantitative analysis. We compute two types of metrics for quantitative comparison on every estimated disparity maps: The percentage of pixels with disparity error larger than $t \in \{ 1.0, 2.0, 5.0 \}$, denoted as $o(t)$; and the average L1 loss, denoted as $avg$. We compute those metrics on the whole disparity sequence except the very first frame and calculate the average as the final performance.

\textbf{Original Dataset.} We first evaluate our method on the same data domain where the models are trained. Same generation parameters are applied according to Sec.~\ref{subsec:dataset}, and the result is shown in Tab.~\ref{tab:trainres}. We can see all the methods can achieve accurate results after the supervised training. Our method does not outperform others but also achieves good accuracy. 

\begin{table}[h]
    \caption{Evaluation Result: Same Configuration to the Training Dataset)}
    \label{tab:trainres}
    \begin{center}
        \begin{tabular}{l|cccc}
            \hline
            Original Dataset & o(1.0) & o(2.0) & o(5.0) & avg \\
            \hline
            Spacetime Stereo & 17.68 & 14.76 & 13.76 & 15.140 \\
            CTD & 24.71 & 11.29 & 6.78 & 2.030 \\
            ActiveStereoNet & 12.06 & 8.41 & 6.24 & 2.391 \\
            FADNet & \textbf{7.32} & 5.47 & 3.25 & \textbf{0.812} \\
            RAFT & 7.76 & \textbf{4.60} & \textbf{2.59} & 0.916 \\
            Ours & 7.93 & 4.89 & 2.81 & 0.955 \\
            \hline
            \hline
            Synthetic Indoor Scenes & o(1.0) & o(2.0) & o(5.0) & avg \\
            \hline
            Spacetime Stereo & 55.37 & 38.88 & 30.38 & 39.222 \\
            CTD & 99.07 & 98.13 & 95.52 & 105.999 \\
            ActiveStereoNet & 77.69 & 66.43 & 53.30 & 50.083 \\
            FADNet & 99.30 & 98.63 & 96.79 & 106.427 \\
            RAFT & 57.73 & 47.59 & 39.84 & 56.417 \\
            Ours & \textbf{48.89} & \textbf{36.39} & \textbf{25.26} & \textbf{24.046} \\
            \hline
            \hline
            Real-World Rigid Motion & o(1.0) & o(2.0) & o(5.0) & avg \\
            \hline
            Spacetime Stereo & 74.86 & 55.23 & 36.88 & 25.772 \\
            CTD & 95.17 & 90.44 & 76.63 & 17.129 \\
            ActiveStereoNet & 76.46 & 57.33 & 32.94 & 9.618 \\
            FADNet & 96.39 & 92.93 & 83.94 & 14.457 \\
            RAFT & 76.01 & 52.84 & 22.60 & 7.666 \\
            Ours & \textbf{73.52} & \textbf{49.65} & \textbf{19.16} & \textbf{5.507} \\
            \hline
        \end{tabular}
    \end{center}
\end{table}

\textbf{Real-world Rigid Motion.} After the evaluation on the Original Dataset, we test the performance on the unseen real-world data. For ground truth generation, we first scan the accurate 3D information of several objects by a precise laser scanner~\cite{freescanx5} offline. Then we place those objects on a rotation plate for data collection. To register the video and scanned model, we use the aruco marker~\cite{garridojurado2016generationof} pasted on the plate to get the accurate position of objects in every frame. We apply a hard mask to remove the background, for we only want to focus on the reconstruction result of the object. Fig.~\ref{fig:realviz} shows the estimated disparity map, and Tab.~\ref{tab:trainres} illustrate the quantitative comparison. Our method outperforms all the SOTA methods with better boundary performance.

\textbf{Synthetic Indoor Scenes.} We also infer our model on another synthetic dataset to test the generalization ability. We use the depth map from ICL-NUIM\cite{ankur2014abenchmark}, a synthetic dataset used for VO/SLAM problems where the camera moves freely inside a room. We use four sequences from a living room for evaluation. We first convert all the RGB images into gray ones, then we create the virtual projector to project the pattern into every image. Different from the original training dataset, these scenes are static and the camera has complex ego-motion with large rotation and transportation. The evaluation result, shown in Tab.~\ref{tab:trainres} and Fig.~\ref{fig:realviz}, illustrate that our method can still get robust results while others suffer a lot from the domain shifting problem.

\begin{table}[h]
    \caption{Computing time comparison. All the methods are evaluated on a GTX1070 card.}
    \label{tab:efficiency}
    \begin{center}
        \begin{tabular}{l|c}
            \hline
             & FPS \\
            \hline
            CTD & 24.0 \\
            ActiveStereoNet & 21.7 \\
            FADNet & 3.5 \\
            RAFT & 16.5 \\
            Ours & \textbf{54.7} \\
            % CTD~\cite{riegler2019connecting} & 24.0 \\
            % ActiveStereoNet~\cite{zhang2018active} & 21.7 \\
            % FADNet~\cite{wang2020afast} & 3.5 \\
            % RAFT~\cite{teed2020raft} & 16.5 \\
            % Ours & \textbf{54.7} \\
            \hline
        \end{tabular}
    \end{center}
\end{table}

\textbf{Efficiency comparison.} We also compare the computing time of disparity estimation, which is shown in Tab.~\ref{tab:efficiency}. The 3D cost volume in ActiveStereoNet~\cite{zhang2018active} and recurrent iteration module in RAFT~\cite{teed2020raft} requires a longer time for evaluation. In our method, we extend the iteration process into the temporal space, resulting in a boost in efficiency. All the results are evaluated on a GTX1070 card.

\begin{table}[h]
    \caption{Ablation Study for Pattern Flow (PF) and Temporal Hidden Layer ($h$).}
    \label{tab:ablation}
    \begin{center}
        \begin{tabular}{l|cccc}
            \hline
            & o(1.0) & o(2.0) & o(5.0) & avg \\
            \hline
            Ours - $h$ - PF & 77.55 & 55.51 & 23.58 & 9.811 \\
            Ours - PF & 76.18 & 51.49 & 20.53 & 6.825 \\
            Ours & \textbf{73.52} & \textbf{49.65} & \textbf{19.16} & \textbf{5.507} \\
            \hline
        \end{tabular}
    \end{center}
\end{table}

\subsection{Ablation Study}
To prove the contribution of the temporal module, we apply the ablation study by removing pattern flow propagation and hidden layer passing. We take the result from Real-World Rigid Motion data for comparison. We first removed the pattern flow warping module(-PF). The previous disparity and hidden features will pass directly to the next frame. By removing the warping, the accuracy decreases. We then further remove the hidden layer(-$h$) and all the network only takes previous disparity as the initial estimation. As listed in Tab.~\ref{tab:ablation}, the accuracy decreased when removing the temporal propagation and pattern flow warping.

We also show several estimated disparity maps in the same sequence of "Synthetic Indoor Scenes" to illustrate the performance improvements along the time. In Fig.\ref{fig:seqviz}, our method can achieve better performance frame by frame in the 30-50 frames from the beginning. Such accurate results can be maintained in a relatively long sequence without any initialization processing. The TIDE-Net takes pattern, a global reference in structured light systems, as the input for every updating process. Therefore our method can be robust to the error accumulation problem.

%% file: chapters/5_Conclusion.tex
\section{Conclusion and Future Works}\label{chap:conclusion}

In this paper, we proposed TIDE-Net, a temporally incremental disparity estimation network for mono-camera structured light systems. The network utilized the temporal coherence for image sequence to improve accuracy and efficiency. With the TIDE-Net's help, disparity maps can be incrementally estimated by focusing on the residual part, resulting in a smaller network size and better generalization ability. We also first introduce and analyze the pattern flow, an active light flow in structured light systems, which could be used to propagate disparity and hidden layers between frames. With the help of pattern flow, the TIDE-Net could focus on the local non-linear part optimization and the accuracy could be further improved. Our method achieves state-of-the-art accuracy on the unseen data. The code of our research will be available on our project page soon.

Our framework also has several limitations for dynamic scene reconstructions. For example, fast-moving objects and complex textures, which are classic challenges in active sensor systems, will harm the correspondence finding in our methods. Besides, our method relies on the detection of pattern moving. Such moving may be very trivial when the baseline is too short and the disparity change is relatively small. In such a situation, our method will degrade to the naive temporal aggregation method.

In our future work, we plan to focus on the model online adaptation research. By updating the network online, the model can adapt to different situations. Therefore the model can perform better on long sequence processing and results can be further improved.